\documentclass{article}

\PassOptionsToPackage{numbers, compress}{natbib}
\usepackage[preprint]{neurips_2026}

\usepackage[utf8]{inputenc} 
\usepackage[T1]{fontenc}    
\usepackage{hyperref}       
\usepackage{url}            
\usepackage{booktabs}       
\usepackage{amsfonts}       
\usepackage{nicefrac}       
\usepackage{microtype}      
\usepackage{xcolor}         
\usepackage{amsmath}
\usepackage{booktabs}
\usepackage{multirow}
\usepackage{graphicx}
\newcommand{\std}[1]{{\scriptsize $\pm$#1}}

\title{CAST: Collapse-Aware multi-Scale Topology Fusion for Multimodal Coreset Selection}

\author{%
  Boran Zhao\thanks{Equal contribution.}\\
  School of Software Engineering,\\
  the National Key Laboratory of Human-Machine Hybrid Augmented Intelligence,\\
  National Engineering Research Center for Visual Information and Applications,\\
  and Institute of Artificial Intelligence and Robotics\\
  Xi'an Jiaotong University\\
  \texttt{boranzhao@xjtu.edu.cn}
  \And
  Hetian Liu\footnotemark[1]\\
  School of Software Engineering\\
  Xi'an Jiaotong University\\
  \texttt{hetianliu@stu.xjtu.edu.cn}
  \And
  Zhenxian Hu\\
  XJTU-POLIMI Joint School\\
  Xi'an Jiaotong University\\
  \texttt{zhenxian.hu@mail.polimi.it}\\
  \And
  Yuqing Yuan\\
  Faculty of Electronic and Information Engineering\\
  Xi'an Jiaotong University\\
  \texttt{2213420425@stu.xjtu.edu.cn}
  \And
  Yu Yan\\
  School of Human Settlements and Civil Engineering\\
  Xi'an Jiaotong University\\
  \texttt{2227402307@stu.xjtu.edu.cn}
  \And
  Pengju Ren\\
  the National Key Laboratory of Human-Machine Hybrid Augmented Intelligence,\\
  National Engineering Research Center for Visual Information and Applications,\\
  and Institute of Artificial Intelligence and Robotics\\
  Xi'an Jiaotong University\\
  \texttt{pengjuren@xjtu.edu.cn}
}

\begin{document}

\maketitle

\begin{abstract}


The training of large multimodal models fundamentally relies on massive image-text datasets, which inevitably incur prohibitive computational overhead. Dataset selection offers a promising paradigm by identifying a highly informative coreset.
However, existing approaches suffer from two critical limitations: (i) single-modality-dominated sampling methods, which ignore the fine-grained \textit{cross-modal information imbalance} inherent in multimodal datasets and thus lead to semantic loss in the other modality; and (ii) coarse-grained sample-scoring-based sampling methods, where the selected coreset tends to be biased toward the scoring model, making it difficult to guarantee distributional equivalence between the coreset and the original dataset. Meanwhile, existing distribution matching and discrete sampling strategies often fail to jointly account for global semantic structure, local fine-grained details, and redundancy-aware coverage in dense regions. 
To this end, we propose CAST, a \textbf{C}ollapse-\textbf{A}ware multi-\textbf{S}cale \textbf{T}opology fusion framework for multimodal coreset selection. We first construct image- and text-modality topologies, and derive a unified topology via \textbf{local-collapse-aware refinement} and \textbf{cross-modal fusion}. We then introduce a \textbf{multi-scale distribution matching} criterion in the diffusion wavelet domain, encouraging the coreset to approximate the original dataset at multiple scales. Finally, we introduce a \textbf{local soft relational coverage} mechanism that extends pure geometric coverage to relation-aware indirect coverage, penalizing redundant selections in dense clusters. 
Extensive experiments on Flickr30K and MS-COCO show that CAST outperforms existing dataset selection baselines, showcasing great superiority in cross-architecture generalization and energy efficiency over state-of-the-art multimodal synthesis methods.

\end{abstract}

\section{Introduction}

The unprecedented success of multimodal large language models (MLLMs) is largely attributed to training on large-scale datasets. However, scaling up datasets incurs prohibitive energy consumption (for instance, training the Gemini~\cite{gemini} large language model required approximately $\mathcal{O}(10^3)$ MWh of energy, enough to power 94 households for an entire year~\cite{adapsne}). Furthermore, prior studies have demonstrated that these massive datasets exhibit substantial redundancy, and only a small fraction of critical samples plays a pivotal role in advancing model capabilities~\cite{truthinfew}. To this end, researchers have proposed dataset compression techniques, which aim to obtain a minimal yet highly informative set of samples to act as a surrogate for the full dataset during training.


Unimodal dataset compression methods include dataset synthesis and selection. Dataset synthesis methods~\cite{datasynthesis_1,datasynthesis_2,datasynthesis_3} typically train specific architectures to generate synthetic samples, which may introduce architecture bias and limit generalization to unseen architectures~\cite{dq}. In contrast, dataset selection methods directly select representative samples from the original dataset without relying on specific architectures, forming coresets with better independence and generalization~\cite{fast}. Therefore, dataset selection has emerged as a promising direction for multimodal dataset compression. However, unlike unimodal datasets, multimodal datasets exhibit several unique characteristics.

\begin{figure}
    \centering
    \includegraphics[width=1\linewidth]{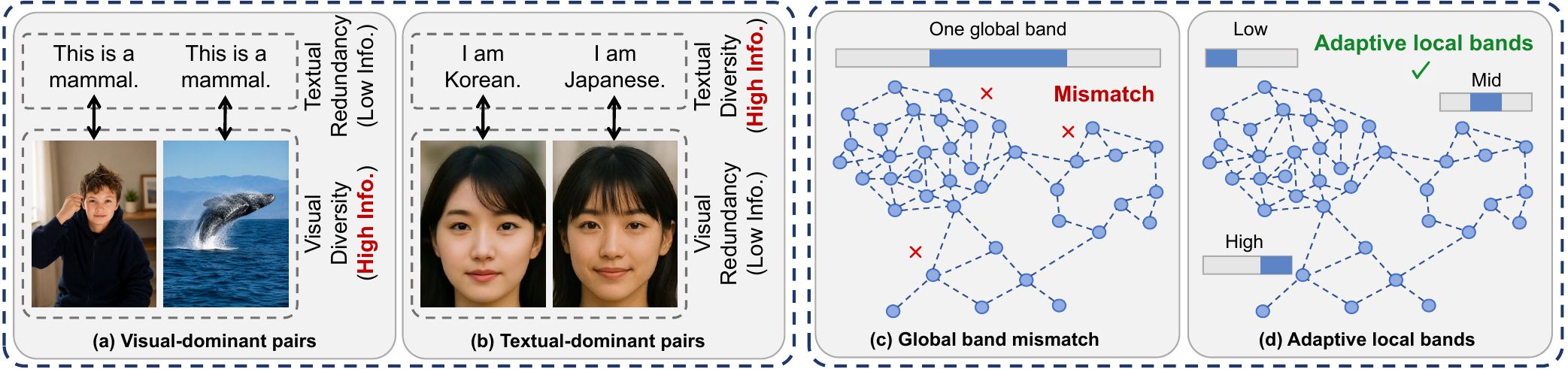}
    \caption{Illustration of \textbf{fine-grained cross-modal information imbalance} and \textbf{spatially varying frequency distributions}. Information density varies drastically across modalities. (a) Visual-dominant pairs with textual redundancy. (b) Text-dominant pairs with visual redundancy. (c) Global spectral band ignores region-specific frequency variations and causes local mismatches. (d) Adaptive local bands capture spatially varying topology by selecting region-specific frequency ranges.}
    \label{fig_imblence_information}
\end{figure}

First, as shown in the left part of Figure~\ref{fig_imblence_information}, multimodal datasets are characterized by \textit{fine-grained information imbalance}. Existing methods often ignore sample-level information content and rely on a single dominant modality, resulting in informative sample loss and limited generalization. For instance, PreSel~\cite{presel} and ViSA~\cite{visa} predominantly anchor their sampling processes on the image modality. However, we observe that the visual modality is not universally dominant; samples retention should be adaptively determined through a holistic evaluation of information from both modalities. To this end, we adopt a topology-graph-based modality information fusion strategy by constructing modality-specific graphs. We first introduce a local-collapse-aware refinement mechanism to alleviate intra-modal fine-grained information imbalance, and then employ a cross-modal fusion mechanism to adaptively adjust the information extraction ratios. In this way, the limitation of fixed single-modality reliance is mitigated, and multimodal graphs are integrated into a unified topology.

Second, the current unimodal state-of-the-art FAST~\cite{fast} transforms the topology graph into the spectral domain and evaluates the distributional discrepancy by selecting global frequencies. Given its identical objective of topology matching, FAST is theoretically applicable to our framework. However, as shown in the right part of Figure~\ref{fig_imblence_information}, we find that in fused multimodal topology, the optimal evaluation frequency bands are location-dependent~\cite{local_frequency_1,local_frequency_2,local_frequency_3}. To overcome this global-scale limitation, we propose a wavelet-domain multi-scale distribution matching criterion that jointly accounts for topological scale and node location, enabling adaptive measurement of the structural discrepancy between the coreset and the original dataset across diverse topological regions. 

Finally, existing methods~\cite{fast,self_fliter} underestimate the ability of proxy points to cover multiple similar samples within the original dataset, resulting in redundant proxy coverage in high-density regions. To tackle this, we introduce the Local Soft Relational Coverage (LSRC) mechanism.


The main contributions of this work are: (1) Local-collapse-aware refinement and cross-modal fusion mechanisms to resolve fine-grained information imbalance; (2) A wavelet-domain multi-scale distribution matching criterion that overcomes location-dependent frequency limitations by preserving global structures and local semantics; and (3) A local soft relational coverage mechanism that softly covers structural neighborhoods, mitigating proxy redundancy and maximizing coreset diversity.

\section{Related Work}

\textbf{Dataset Compression.} Dataset compression is generally classified into dataset synthesis and dataset selection~\cite{DDsurvey}. Dataset synthesis~\cite{datasynthesis_1, datasynthesis_2, datasynthesis_3} aims to distill the information of original datasets into artificially generated samples. Existing methods typically synthesize such samples by matching training parameters~\cite{parameter}, gradients~\cite{gradientmatch_1,gradientmatch_2}, or trajectories~\cite{trajectory_1,trajectory_2} between real and synthetic data.
Recently, some efforts have extended to multimodal scenarios. 
For instance, LoRS~\cite{lors} overlooks intra-modal structures, leading to modality collapse. While RepBlend~\cite{repblend} mitigates this via representation blending, it misattributes the collapse to the compression process. Instead, we argue that modality collapse stems from inherent cross-modal information imbalance: samples indistinguishable in one modality may remain distinct in another. Furthermore, dataset synthesis paradigms inherently suffer from prohibitive computational overhead and inevitably sacrifice the fine-grained semantics and structural details of authentic samples. In contrast, dataset selection directly selects real samples from the original dataset to form a coreset, avoiding semantic distortion from synthesized samples and additional generation costs.

\textbf{Multimodal Dataset Selection.} Multimodal dataset selection mainly falls into two categories. The first adopts the image modality as the primary anchor for sampling. For instance, PreSel~\cite{presel} and ViSA~\cite{visa} filter redundancy based on clustering in the visual feature space and visual complexity, respectively. However, this unimodal strategy ignores inherent cross-modal imbalances, risking the loss of complementary information and weakening coreset representativeness.
To incorporate information from different modalities, another line of methods evaluates samples through coarse-grained scoring. For instance, RAP~\cite{truthinfew} filters samples according to output-level reasoning discrepancies, while DataProphet~\cite{dataprophet} ranks dataset value using metrics based on multimodal perplexity, similarity, and diversity. Self-filter~\cite{self_fliter} uses the Vision-Language Model (VLM) itself as a filter and trains a scoring network for sample evaluation and selection. However, such scoring-centric methods are susceptible to the scoring model or heuristic metrics, making it difficult to ensure distributional equivalence between the coreset and the original dataset.


These limitations indicate that multimodal dataset selection requires a unified representation that integrates visual and textual semantics. Graph topologies, by modeling pairwise similarity relations, are well-suited to capture the distribution of the original dataset. While graph topologies naturally capture macroscopic dataset distributions, existing "graph abstraction–graph fusion" methods~\cite{graphfusion_1,graphfusion_2} operate under the idealized premise that independently constructed unimodal graphs possess high structural reliability. This ignores fine-grained multimodal imbalances, causing unrefined modality-specific graphs to propagate unreliable relations during fusion. To address this, we propose an cross-modal local-collapse-aware refinement and cross-modal fusion mechanism, correcting locally degraded structures prior to region-adaptive fusion.




\textbf{Distribution Matching and Coreset Selection.} Coreset quality hinges on robust distribution matching and effective sampling strategies. Traditional Euclidean (e.g., MSE~\cite{mse}, MMD~\cite{mmd}) or probabilistic (e.g., KL~\cite{kl}) metrics fail to capture high-order structural moments. While FAST~\cite{fast} introduces global frequency-domain alignment to address this, its macroscopic approach overlooks location-dependent discrepancies in multimodal topologies. Consequently, we propose a wavelet-domain multi-scale matching criterion to explicitly inject topological locality into frequency selection. Furthermore, existing methods~\cite{fast,self_fliter} typically adopt independent hard sampling mechanisms. This rigid one-to-one mapping ignores a proxy's ability to indirectly cover its neighborhood. To address this, our local soft relational coverage mechanism enables proxies to softly cover locally similar samples, significantly improving the coreset's information density and diversity.



\section{Method}

\subsection{Overview}



\begin{figure}
    \centering
    \includegraphics[width=1\linewidth]{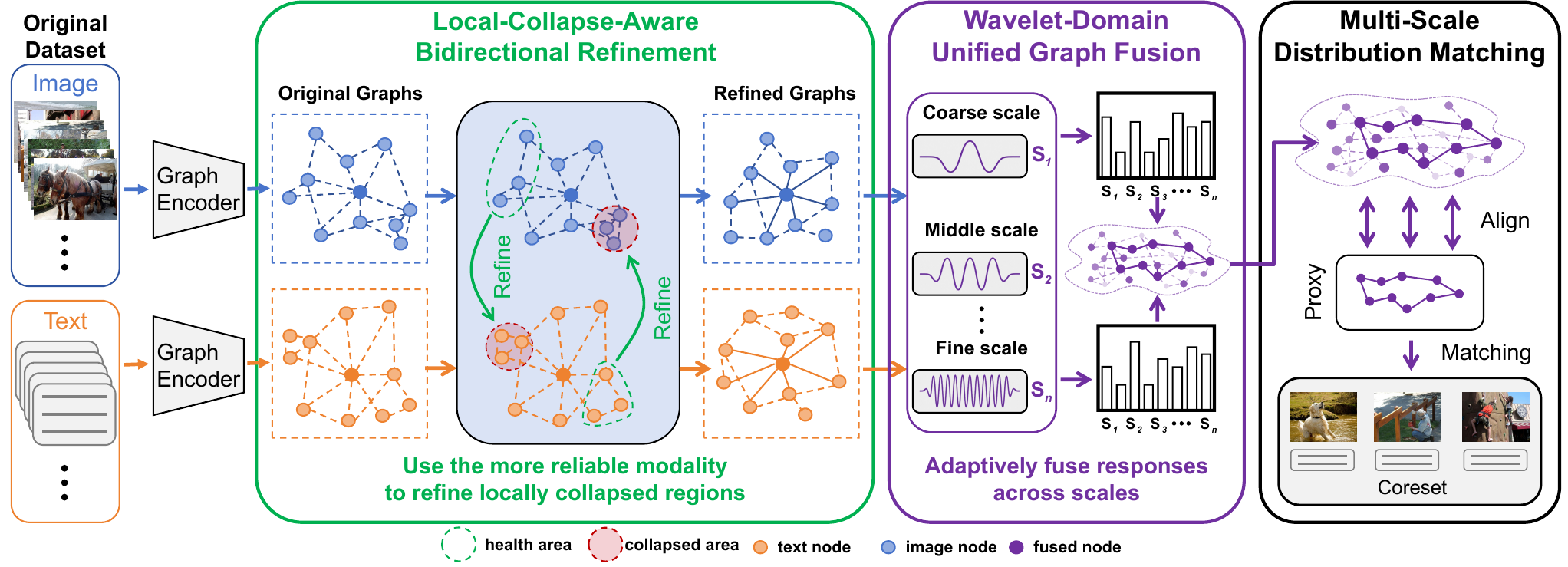}
    \caption{The overall pipeline of the proposed CAST framework.}
    \label{fig_main}
\end{figure}

Given a multimodal dataset $\mathcal{D}=\{x_i\}_{i=1}^N$, where $x_i=(x_i^{(I)},x_i^{(T)})$, our objective is to select a compact coreset $\mathcal{C} \subset \mathcal{D}$ with $|\mathcal{C}|=K$, such that $\mathcal{C}$ preserves the structural distribution of $\mathcal{D}$ across modalities. To this end, we propose CAST, a Collapse-Aware multi-Scale Topology fusion framework for multimodal coreset selection. As illustrated in Figure~\ref{fig_main}, our methodology is systematically carried out in four stages. First, we construct original graphs via Graph Encoder. Second, we derive a unified topology via a local-collapse-aware bidirectional refinement and wavelet-domain unified fusion mechanism. Third, anchored on the fused topology, we formulate a multi-scale distribution matching criterion, ensuring that the proxy maintains consistency with the original dataset. Finally, we cast the coreset selection process as a local soft relational coverage problem, empowering each selected proxy to cover its local topological neighborhood.

\subsection{Graph Construction}



To capture the intra-modal neighborhood similarity, we construct unimodal graphs for both modalities. Specifically, we first extract modality-specific feature representations for each sample, denoted as $z_i^{(I)}$ and $z_i^{(T)}$. Notably, the feature extractors only serve as a foundational representation provider, rather than surrogate models co-optimized.
Subsequently, we construct a local $k$-nearest neighbor ($k$-NN) graph based on the fuzzy topology theory of UMAP~\cite{umap}. For a given node $i$ within modality $m \in \{I, T\}$, the local scale parameter $\sigma_i^{(m)}$ is derived by solving the following constraint:
$$\sum_{j=1}^{k}\exp\left(-\frac{\max(0,\,d^{(m)}(x_i,x_j)-\rho_i^{(m)})}{\sigma_i^{(m)}}\right)=\log_2 k$$
where $d^{(m)}(x_i,x_j)$ denotes the distance in the feature space of modality $m$. The term $\rho_i^{(m)}$ is determined by the distance to the nearest neighbor.
Since the underlying manifold topology should be represented by undirected relations, we further symmetrize the directed edges and construct the unimodal topology matrix as $B_{ij}^{(m)}=A_{i\rightarrow j}^{(m)}+A_{j\rightarrow i}^{(m)}-A_{i\rightarrow j}^{(m)}A_{j\rightarrow i}^{(m)}$.
Ultimately, we derive the visual topological graph $B^{(I)}$ and the textual topological graph $B^{(T)}$, which serve as the foundational inputs for the subsequent cross-modal refinement phase.

\subsection{Collapse-aware Refinement and Fusion}


\textbf{Cross-modal Local-Collapse-Aware Bidirectional Refinement.}
As discussed earlier, multimodal data often exhibit fine-grained information imbalance, which leads to local collapsed structures within both $B^{(I)}$ and $B^{(T)}$. Therefore, we introduce an cross-modal bidirectional local-collapse-aware refinement mechanism before constructing the unified graph. Specifically, we quantify the collapse severity using the local relation redundancy $R_i^{(m)}$ and the neighborhood inseparability $S_i^{(m)}$, and then use the healthy local topology of one modality to refine the structure of the other.


Specifically, for node $i$ within modality $m \in \{I, T\}$, we characterize its local structural degradation from two complementary perspectives: local relation redundancy $R_i^{(m)}$ and neighborhood inseparability $S_i^{(m)}$. The former measures whether the local patterns of node $i$ and its neighbors $\mathcal{N}(i)$ are overly similar, with a larger $R_i^{(m)}$ signifying a more homogenized connectivity landscape in modality $m$. The latter evaluates the degradation of discriminability; a larger $S_i^{(m)}$ implies a flattened edge weight distribution towards $\mathcal{N}(i)$, indicating a local collapse that blurs fine-grained semantic boundaries among neighbors. Formally, let $b_i^{(m)}$ denote the edge weight vector of node $i$ over the unified candidate neighborhood, these two metrics are formulated as:
$$R_i^{(m)}=\frac{1}{|\mathcal{N}(i)|}\sum_{j\in\mathcal{N}(i)}\text{sim}\left(b_i^{(m)},b_j^{(m)}\right), \quad S_i^{(m)}=-\frac{1}{\log|\mathcal{N}(i)|}\sum_{j\in\mathcal{N}(i)}p_{ij}^{(m)}\log(p_{ij}^{(m)}+\epsilon)$$
where $\text{sim}(\cdot,\cdot)$ represents the similarity (e.g, cosine similarity) between the local edge weight patterns of two nodes, and $p_{ij}^{(m)}$ represents the local edge weight distribution obtained by normalizing the raw weights $B_{ij}^{(m)}$ across all neighbors of node $i$ under modality $m$. Intuitively, if this distribution approaches uniformity, distinguishing fine-grained structural subtleties becomes exceedingly difficult, which is precisely reflected by a larger $S_i^{(m)}$.

Synthesizing $R_i^{(m)}$ and $S_i^{(m)}$, we assess node-level structural degeneration to detect local collapse. This degradation is then propagated to estimate the reliability of each edge. For cross-modal refinement, edges with significantly lower reliability in one modality receive bounded compensation from the other. This selective constraint prevents excessive topological distortion. Consequently, the raw graphs $B^{(I)}$ and $B^{(T)}$ are refined into $\hat{B}^{(I)}$ and $\hat{B}^{(T)}$ for subsequent unified fusion.




\textbf{Diffusion-Wavelet-Based Multi-scale Latent Topology Fusion.}
To simultaneously preserve the global semantic backbone and local fine-grained structures of multimodal data, we propose a multi-scale latent topology fusion method in the diffusion wavelet domain. Specifically, we first project the refined graphs $\hat{B}^{(I)}$ and $\hat{B}^{(T)}$ into a multi-scale diffusion response space, and evaluate the reliability of the two modalities at different topological scales $s$. Here, the scale $s$ denotes the step size of the diffusion process, and also corresponds to the observation granularity of the topology. Small scales focus on local neighborhoods and high-frequency boundaries, while large scales capture the global semantic skeletons after long-range diffusion. Guided by these scale-dependent reliabilities, we adaptively fuse the bimodal responses and further reconstruct a unified latent topology graph $B^*$.


Specifically, we first derive the random walk transition matrix $P^{(m)}=(D^{(m)})^{-1}\hat{B}^{(m)}$ for the refined graph $\hat{B}^{(m)}$, where $D^{(m)}$ is the degree matrix associated with $\hat{B}^{(m)}$. Subsequently, the diffusion wavelet response at scale $s$ is defined as $W_s^{(m)} = (P^{(m)})^sQ-(P^{(m)})^{2s}Q$. Here, $Q$ serves as a probe signal used to observe the responses of the image/text graphs at different scales. To determine which modality provides more reliable structural information at different scales and adaptively allocate fusion weights, we formulate the response entropy for modality $m$ at scale $s$ as follows:$$H_s^{(m)}=-\frac{1}{\log N}\sum_i\pi_{s,i}^{(m)}\log(\pi_{s,i}^{(m)}+\epsilon)$$where the term $\pi_{s,i}^{(m)}=\frac{\|W_{s,i}^{(m)}\|_2^2}{\sum_u \|W_{s,u}^{(m)}\|_2^2+\epsilon}$ measures the normalized response energy of node $i$ under scale $s$ within modality $m$.


A higher $H_s^{(m)}$ indicates a balanced response distribution across nodes, whereas a lower value implies that the structural representation at that scale may be dominated by localized regions. Guided by entropy-derived weights, we fuse the visual and textual diffusion wavelet responses to derive the target consensus response at each scale. Finally, by assessing pairwise similarities within this consensus, we inversely reconstruct the unified graph $B^*$ over the joint candidate edge support set of the refined bimodal graphs (details deferred to Appendix~\ref{appendix_fusion}).

\begin{figure}
    \centering
    \includegraphics[width=1\linewidth]{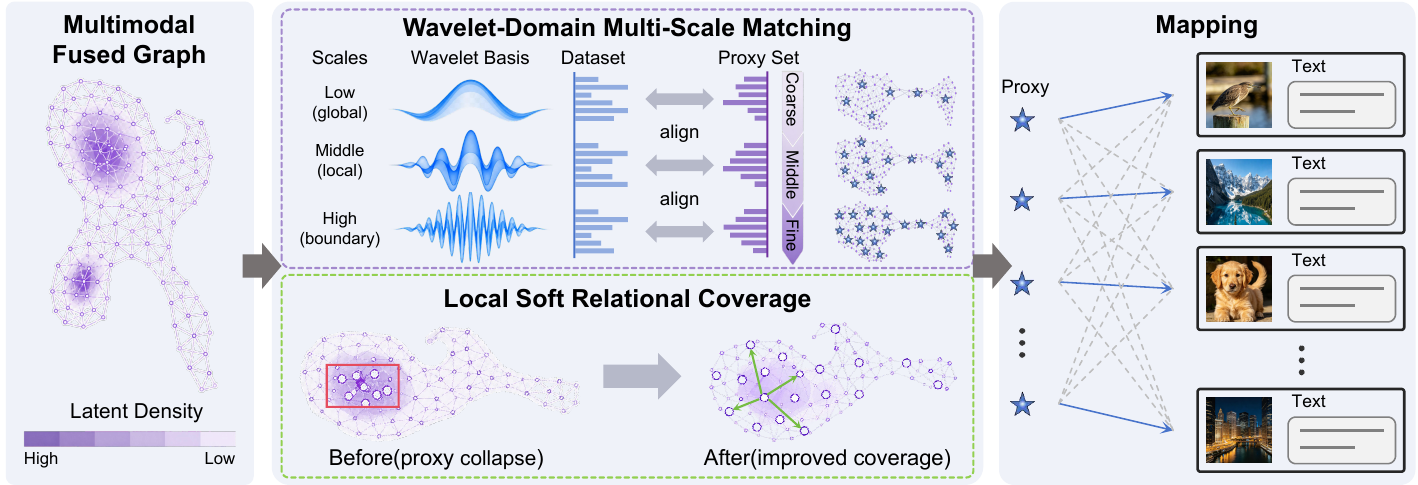}
    \caption{Illustration of the coreset selection optimization in CAST.}
    \label{fig_matching_LSRC}
\end{figure}

\subsection{Wavelet-Domain Multi-scale Matching}


Based on the unified latent graph $B^*$, we optimize a set of continuous proxy points $Y=\{y_1,y_2,\ldots,y_K\}$ and then map these proxy points back to real samples to construct the coreset $\mathcal{C}\subset \mathcal{D}$ with $|\mathcal{C}|=K$. As illustrated in Figure~\ref{fig_matching_LSRC}, to compel these proxies to faithfully approximate the distribution of the original dataset across varying scales, we jointly compute three matching losses: distribution alignment, edge protection, and frequency-domain coverage.


To formulate the optimization objectives, we first define the diffusion wavelet response: $\Phi_s(Z)=(P^*)^sZ-(P^*)^{2s}Z$ where $P^*$ denotes the random walk transition matrix constructed on $B^*$. The term $Z$ represents the fused representations, derived by concatenating the normalized image features $z_i^{(I)}$, text features $z_i^{(T)}$, and the diffusion embeddings on $B^*$, followed by a final normalization. Correspondingly, the diffusion wavelet response of a proxy point $y_k$ at scale $s$ is parameterized as a soft interpolation of its topological neighbors:
$$\Phi_s(y_k)=\sum_{i\in\mathcal{N}(y_k)}\eta_{ki}\Phi_s(z_i),\qquad \sum_{i\in\mathcal{N}(y_k)}\eta_{ki}=1$$
where the term $\eta_{ki}$ serves as a distance-aware soft interpolation weight, effectively computed via a normalized exponential decay based on the proximity between $y_k$ and its neighbors.


To faithfully represent the original dataset, our matching strategy is primarily driven by a \textbf{global distribution alignment loss}, supplemented by two auxiliary objectives: an \textbf{edge protection loss} and a \textbf{frequency-domain coverage loss}. Specifically, the primary loss enforces overall consistency by minimizing the Sliced Wasserstein Distance (SWD) between the diffusion wavelet responses. Concurrently, to prevent high-frequency details from being smoothed out, the edge loss forces proxies to cover boundaries by penalizing the distance $\delta_i^{(s)}=\min_k\|\Phi_s(z_i)-\Phi_s(y_k)\|_2^2$ for nodes with high inherent response energy $e_i^{(s)}=\|\Phi_s(z_i)\|_2^2$. Meanwhile, to explicitly prevent sparse regions from being overlooked, the auxiliary coverage loss evaluates the negative log-likelihood of the proxy-to-node coverage degree $\operatorname{cov}_i^{(s)}=(1/K)\sum_{k=1}^{K}\exp\left(-\|\Phi_s(z_i)-\Phi_s(y_k)\|_2^2/\tau\right)$. These three concurrent objectives at scale $s$ are succinctly formulated as:
$$L_{\text{dist}}^{(s)}=\operatorname{SWD}\bigl(\Phi_s(Z),\Phi_s(Y)\bigr), \quad L_{\text{edge}}^{(s)}=\frac{\sum_i e_i^{(s)}\delta_i^{(s)}}{\sum_i e_i^{(s)}+\epsilon}, \quad L_{\text{cov}}^{(s)}=-\frac{1}{N}\sum_i\log(\operatorname{cov}_i^{(s)}+\epsilon)$$

By integrating them, we employ a coarse-to-fine scale activation strategy to formulate the total matching objective in the wavelet domain:
$$L_{\text{wavelet}} = \sum_{s\in\mathcal{S}_t}\beta_s\left[L_{\text{dist}}^{(s)}+\lambda_{\text{edge}}L_{\text{edge}}^{(s)} + \lambda_{\text{cov}}L_{\text{cov}}^{(s)}\right]$$
This strategy drives the proxy points to capture the global macro-distribution at coarse scales, while progressively unlocking finer scales to firmly anchor near regions with drastic structural changes and complete frequency coverage.

\subsection{Local Soft Relational Coverage}


While multi-scale distribution matching effectively approximates the dataset, relying solely on geometric distance may causes redundant proxy assignments in dense local regions. To mitigate this, we introduce the Local Soft Relational Coverage (LSRC) module, which extends point-to-point geometric coverage to a relation-aware indirect paradigm.

Specifically, we extend the coverage mechanism to the spatial domain. By introducing the soft relation graph $R$—which explicitly fuses spatial geometric proximity with cross-modal topological priors—we propagate the basic spatial direct coverage $h_i$ of proxy points on node $z_i$ (calculated in the same form as $\operatorname{cov}_i^{(s)}$, but based on the geometric distance $\|z_i-y_k\|_2^2$ in the underlying spatial domain) through relation-aware propagation, forming the relation-aware indirect coverage $\bar{h}_i$.
Therefore, proxy points not only cover their current geometric space but can also represent their neighboring nodes based on topological priors, generating a strong local coverage repulsion. The derived LSRC loss (which unifies the indirect coverage sufficiency and the local relation smoothness constraint) is simplified as follows:$$L_{\text{LSRC}} = -\frac{1}{N}\sum_i\log(\bar{h}_i+\epsilon) + \mu \frac{\sum_{(i,j)}R_{ij}(h_i-h_j)^2}{\sum_{(i,j)}R_{ij}+\epsilon}$$Finally, by combining the scale-dependent wavelet matching with the scale-independent spatial coverage, the overall objective for continuous proxy optimization is integrated as:$$L = L_{\text{wavelet}} + \lambda_{\text{LSRC}}L_{\text{LSRC}} + \lambda_{\text{reg}}L_{\text{reg}}$$where $L_{\text{reg}}$ contains auxiliary regularization terms such as anti-collapse and diversity constraints. Jointly governed by this unified objective, the proxy points not only seamlessly maintain multi-scale consistency with the raw dataset but also actively circumvent redundant resource allocation in dense regions. Upon convergence, following FAST~\cite{fast}, we employ the Hungarian algorithm to map the optimized continuous proxy points back to authentic image-text pairs, thereby forming the final coreset $\mathcal{C}\subset\mathcal{D}$ (details deferred to Appendix~\ref{appendix_LSRC_and_matching}).

\section{Experiments}
\subsection{Experimental Setup}


\textbf{Datasets and Metrics.} Following~\cite{lors} and~\cite{repblend}, we evaluate our method on image captioning datasets: Flickr30K~\cite{flickr} and MS-COCO~\cite{mscoco}. We adopt the Karpathy split~\cite{split}, where Flickr30K contains 29k/1k/1k images and MS-COCO contains 113k/5k/5k images for training/validation/test. Each image is paired with five human-annotated captions. We evaluate bidirectional retrieval performance using Recall@K with $K \in \{1, 5, 10\}$. Specifically, IR@K denotes text-to-image retrieval, and TR@K denotes image-to-text retrieval.


\textbf{Baselines.}
We compare our method with three groups of baselines. \textbf{(1) unimodal dataset selection methods:} Random, Herding~\cite{herding}, Glister~\cite{glister}, EL2N~\cite{el2n}, Entropy~\cite{entropy}, GradMatch~\cite{gradmatch}, and FAST~\cite{fast}. Since these methods are not originally designed for image-text pairs, we adapt them by constructing a fused multimodal representation from normalized image and text features. \textbf{(2) multimodal dataset selection methods:} ViSA~\cite{visa}, PreSel~\cite{presel}, Dyn-Prune~\cite{dyn_prune}, and DataProphet~\cite{dataprophet}. Since these methods are mainly developed for multimodal instruction selection, we adapt them to image-text pair selection while preserving their original scoring or filtering principles. \textbf{(3) multimodal dataset synthesis methods:} LoRS~\cite{lors} and RepBlend~\cite{repblend}. Since these methods generate synthetic image-text surrogates rather than selecting real samples, we mainly compare with them in cross-architecture generalization and efficiency analysis.

\textbf{Implementation Details.} 
We extract multimodal features using geometric distances at the pixel-level for images and a frozen BERT~\cite{bert} for text. For CAST configurations, the topology is modeled via a $k$-NN graph ($k=15$) constructed based on the common neighbors of nodes. The diffusion wavelet scales are defined as $\mathcal{S}_t=\{1, 2, 4\}$. All experiments are conducted on NVIDIA RTX A6000 GPU.

\subsection{Main Results on Image-Text Retrieval}

\begin{figure}
    \centering
    \includegraphics[width=1\linewidth]{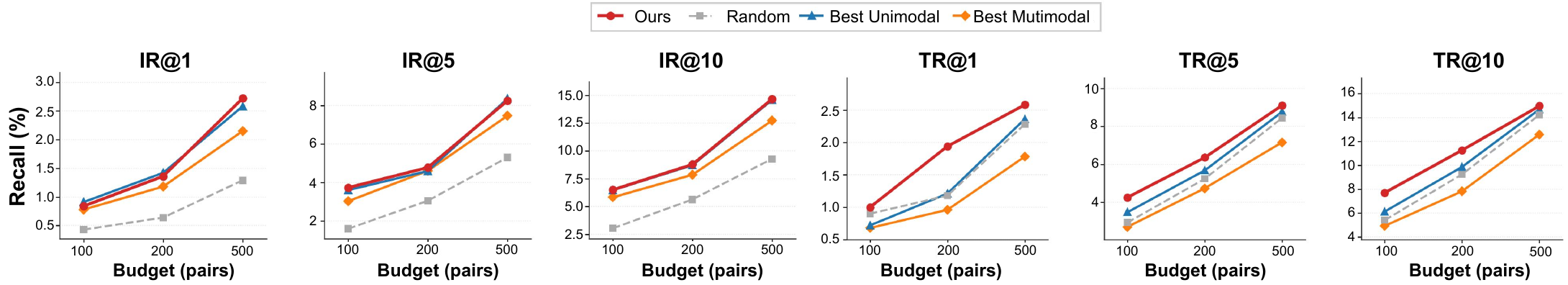}
    \caption{Quantitative results of bidirectional image-text retrieval on the MS-COCO dataset across varying compression budgets, detailed results are provided in Appendix~\ref{appendix_COCO_result}.}
    \label{fig_mscoco_result}
\end{figure}

We evaluate bidirectional image-text retrieval under data budgets of 100, 200, and 500 pairs, summarizing Flickr30K results in Table~\ref{tab_flickr} and MS-COCO trends in Figure~\ref{fig_mscoco_result}. CAST consistently achieves the best performance on Flickr30K and obtains highly competitive results on MS-COCO, especially dominating all text retrieval metrics.
Naively adapted unimodal selection methods yield sub-optimal results because they treat concatenated multimodal features as a flat space, ignoring the inherent structural discrepancies between modalities. In contrast, by introducing a collapse-aware bidirectional refinement and fusion mechanism, our method prevents modality-specific structural degradation from compromising the final coreset. Furthermore, multimodal baselines (e.g., ViSA~\cite{visa}, PreSel~\cite{presel}) generally underperform because their unimodal-dominant sampling strategies fail to capture the global joint distribution of multimodal data, leading to dense-region redundancy and long-tail omissions. To overcome this, by leveraging multi-scale spectral topology and LSRC, CAST circumvents such redundancy and firmly anchors continuous proxies to critical semantic boundaries.

\begin{table*}[t]
    \centering
    \caption{Results of bidirectional image-text retrieval on the Flickr30K dataset across varying compression budgets. The best performance is highlighted in \textbf{bold}.}
    \label{tab_flickr}
    \resizebox{\textwidth}{!}{
    \setlength{\tabcolsep}{3pt}
    \begin{tabular}{cc | ccccccc | cccc | c}
    \toprule
    \multirow{2}{*}{\textbf{Budget}} & \multirow{2}{*}{\textbf{Metric}} & \multicolumn{7}{c|}{\textbf{Unimodal Selection Baselines}} & \multicolumn{4}{c|}{\textbf{Multimodal Selection Baselines}} & \multirow{2}{*}{\textbf{Ours}} \\
    \cmidrule{3-13}
    & & Random & Herding & Glister & EL2N & Entropy & GradMatch & FAST & ViSA & PreSel & Dyn-Prune & Prophet & \\
    \midrule
    \multirow{6}{*}{100} 
    & IR@1  & 1.20 \std{0.14} & 1.20 \std{0.15} & 1.40 \std{0.18} & 1.00 \std{0.12} & 0.90 \std{0.10} & 1.60 \std{0.17} & 1.60 \std{0.19} & 0.40 \std{0.12} & 0.50 \std{0.08} & 1.00 \std{0.15} & 0.80 \std{0.11} & \textbf{2.88} \std{0.74} \\
    & IR@5  & 6.20 \std{0.52} & 5.60 \std{0.48} & 5.80 \std{0.55} & 5.00 \std{0.42} & 2.40 \std{0.28} & 4.70 \std{0.39} & 5.80 \std{0.51} & 1.70 \std{0.21} & 2.40 \std{0.28} & 4.10 \std{0.35} & 2.60 \std{0.22} & \textbf{9.52} \std{1.07} \\
    & IR@10 & 10.50 \std{0.85} & 10.20 \std{0.78} & 8.90 \std{0.65} & 9.00 \std{0.71} & 3.90 \std{0.36} & 7.90 \std{0.62} & 9.80 \std{0.88} & 3.40 \std{0.33} & 4.20 \std{0.38} & 6.50 \std{0.46} & 4.70 \std{0.29} & \textbf{14.98} \std{0.94} \\
    & TR@1  & 1.18 \std{0.15} & 0.48 \std{0.07} & 0.84 \std{0.11} & 0.58 \std{0.08} & 0.50 \std{0.06} & 0.80 \std{0.12} & 1.00 \std{0.13} & 0.52 \std{0.10} & 0.62 \std{0.12} & 0.70 \std{0.14} & 0.44 \std{0.09} & \textbf{1.50} \std{0.08} \\
    & TR@5  & 4.20 \std{0.38} & 2.34 \std{0.25} & 3.42 \std{0.31} & 2.80 \std{0.29} & 1.82 \std{0.21} & 2.76 \std{0.28} & 4.04 \std{0.41} & 2.44 \std{0.25} & 2.10 \std{0.18} & 2.38 \std{0.22} & 2.44 \std{0.27} & \textbf{6.15} \std{0.57} \\
    & TR@10 & 7.48 \std{0.65} & 4.84 \std{0.42} & 5.56 \std{0.48} & 5.36 \std{0.51} & 3.48 \std{0.33} & 4.58 \std{0.44} & 7.02 \std{0.59} & 3.96 \std{0.31} & 3.62 \std{0.29} & 4.56 \std{0.37} & 4.12 \std{0.34} & \textbf{10.45} \std{0.94} \\
    \midrule
    \multirow{6}{*}{200} 
    & IR@1  & 3.20 \std{0.35} & 2.00 \std{0.22} & 2.30 \std{0.26} & 2.20 \std{0.24} & 1.20 \std{0.14} & 2.50 \std{0.28} & 2.10 \std{0.21} & 1.00 \std{0.16} & 0.90 \std{0.14} & 1.50 \std{0.18} & 1.20 \std{0.15} & \textbf{3.83} \std{0.59} \\
    & IR@5  & 10.20 \std{0.88} & 8.20 \std{0.75} & 8.30 \std{0.68} & 6.60 \std{0.59} & 4.20 \std{0.41} & 8.10 \std{0.72} & 10.30 \std{0.91} & 3.20 \std{0.28} & 4.00 \std{0.31} & 5.60 \std{0.42} & 4.50 \std{0.33} & \textbf{12.13} \std{0.72} \\
    & IR@10 & 15.50 \std{1.15} & 14.00 \std{1.22} & 13.20 \std{1.08} & 11.10 \std{0.95} & 6.00 \std{0.55} & 13.60 \std{1.12} & 16.80 \std{1.35} & 6.60 \std{0.45} & 7.20 \std{0.41} & 9.20 \std{0.53} & 7.60 \std{0.48} & \textbf{19.10} \std{0.77} \\
    & TR@1  & 1.66 \std{0.18} & 0.84 \std{0.11} & 0.98 \std{0.12} & 0.98 \std{0.10} & 0.84 \std{0.12} & 1.20 \std{0.15} & 1.76 \std{0.21} & 0.80 \std{0.13} & 0.74 \std{0.11} & 1.00 \std{0.15} & 0.92 \std{0.14} & \textbf{2.05} \std{0.23} \\
    & TR@5  & 6.94 \std{0.61} & 4.06 \std{0.38} & 4.68 \std{0.42} & 4.16 \std{0.35} & 3.06 \std{0.29} & 4.54 \std{0.41} & 7.12 \std{0.68} & 3.42 \std{0.27} & 2.54 \std{0.22} & 3.84 \std{0.31} & 3.72 \std{0.28} & \textbf{7.64} \std{0.59} \\
    & TR@10 & 11.40 \std{0.92} & 7.14 \std{0.65} & 7.82 \std{0.71} & 7.44 \std{0.68} & 5.30 \std{0.45} & 8.04 \std{0.76} & 12.04 \std{1.05} & 5.74 \std{0.39} & 5.08 \std{0.33} & 6.78 \std{0.44} & 6.18 \std{0.36} & \textbf{13.15} \std{0.91} \\
    \midrule
    \multirow{6}{*}{500} 
    & IR@1  & 5.00 \std{0.45} & 3.50 \std{0.31} & 3.50 \std{0.38} & 2.90 \std{0.26} & 2.50 \std{0.22} & 4.50 \std{0.41} & 5.00 \std{0.52} & 2.70 \std{0.22} & 2.80 \std{0.25} & 3.00 \std{0.27} & 2.30 \std{0.19} & \textbf{5.60} \std{1.03} \\
    & IR@5  & 14.80 \std{1.12} & 15.10 \std{1.25} & 14.10 \std{1.18} & 13.20 \std{1.05} & 8.70 \std{0.75} & 14.10 \std{1.21} & 15.60 \std{1.32} & 8.40 \std{0.46} & 7.20 \std{0.42} & 7.70 \std{0.41} & 7.90 \std{0.43} & \textbf{17.43} \std{0.81} \\
    & IR@10 & 23.50 \std{1.65} & 23.30 \std{1.58} & 22.60 \std{1.42} & 20.40 \std{1.35} & 14.20 \std{1.15} & 23.30 \std{1.55} & 23.10 \std{1.14} & 12.30 \std{0.62} & 11.10 \std{0.55} & 13.80 \std{0.68} & 12.10 \std{0.58} & \textbf{26.78} \std{0.60} \\
    & TR@1  & 3.12 \std{0.28} & 2.70 \std{0.25} & 2.44 \std{0.21} & 2.18 \std{0.18} & 1.92 \std{0.16} & 2.38 \std{0.22} & 3.08 \std{0.29} & 1.72 \std{0.18} & 1.74 \std{0.21} & 1.50 \std{0.16} & 1.64 \std{0.17} & \textbf{3.21} \std{0.25} \\
    & TR@5  & 9.76 \std{0.85} & 9.22 \std{0.78} & 9.44 \std{0.82} & 8.84 \std{0.75} & 6.38 \std{0.55} & 9.08 \std{0.81} & 11.32 \std{0.95} & 5.58 \std{0.35} & 5.82 \std{0.38} & 6.38 \std{0.33} & 6.32 \std{0.36} & \textbf{11.67} \std{0.44} \\
    & TR@10 & 16.70 \std{1.25} & 15.42 \std{1.18} & 14.88 \std{1.12} & 14.30 \std{1.05} & 10.44 \std{0.85} & 15.10 \std{1.22} & 17.62 \std{1.38} & 9.16 \std{0.47} & 9.60 \std{0.52} & 10.68 \std{0.55} & 10.42 \std{0.49} & \textbf{19.37} \std{0.79} \\
    \bottomrule
    \end{tabular}
    }
\end{table*}

\subsection{Generalization and Efficiency}

\textbf{Cross-Architecture Generalization.}
To assess coreset robustness against architecture bias, we conduct cross-architecture experiments on Flickr30K under a 3\% compression budget. In our setup, CAST utilizes decoupled raw-pixel and frozen-BERT features, whereas the baselines rely on NFNet~\cite{nfnet} for image features and BERT for text features. We evaluate CLIP-style~\cite{CLIP} dual encoders with three backbone combinations: ViT~\cite{ViT}+BERT, ResNet50~\cite{resnet50}+BERT, and NFNet+BERT. 
As illustrated in Figure~\ref{fig_ablation_general_energy}~(a), baselines severely overfit to their upstream CNN priors, leading to a substantial degradation on the Transformer-based ViT. In contrast, CAST's architecture-agnostic selection better preserves global structural information. This robust unbiased topology allows high-capacity models like ViT to better exploit their representation capacity. Appendix~\ref{appendix_synthesis_cross_architecture} results further show that replacing the NFNet extractor in synthesis methods with a weaker ResNet10~\cite{resnet50} causes their performance to collapse, while CAST remains robust to such capacity bottlenecks.

\textbf{Multimodal Instruction Fine-tuning.}
Beyond discriminative retrieval tasks, we further evaluate the cross-task transferability of CAST on generative multimodal instruction fine-tuning. Specifically, we use the LLaVA-1.5-mix-665K~\cite{llava_1.5_dataset} dataset to fine-tune the Qwen2-VL-2B architecture under budgets of 1\%, 5\%, and 10\%, comparing CAST-selected subsets against the Random baseline. The fine-tuned models are evaluated on a comprehensive suite of multimodal reasoning benchmarks, including GQA~\cite{gqa}, ScienceQA~\cite{scienceqa}, MMBench~\cite{mmbench}, TextVQA~\cite{textvqa}, and POPE~\cite{pope}. Across all compression ratios, CAST consistently outperforms Random, achieving an average improvement of 1--2 points across benchmarks. As shown in Table~\ref{tab_finetune_flat}, results show that CAST does not merely benefit retrieval matching, but also preserves dense and high-quality multimodal semantic structures required for complex visual question answering and generative reasoning.

\begin{table*}[t]
    \centering
    \caption{Performance comparison between our method and random sampling under different subset ratios across multiple benchmarks. The best results for each budget are highlighted in \textbf{bold}.}
    \label{tab_finetune_flat}
    \resizebox{0.75\textwidth}{!}{
    \begin{tabular}{l | cc | cc | cc | cc | cc}
    \toprule
    \multirow{2}{*}{\textbf{Ratio}} & \multicolumn{2}{c|}{\textbf{GQA}} & \multicolumn{2}{c|}{\textbf{ScienceQA}} & \multicolumn{2}{c|}{\textbf{MMBench}} & \multicolumn{2}{c|}{\textbf{TextVQA}} & \multicolumn{2}{c}{\textbf{POPE}} \\
    \cmidrule{2-11}
    & Ours & Rand & Ours & Rand & Ours & Rand & Ours & Rand & Ours & Rand \\
    \midrule
    1\%  & \textbf{59.71} & 58.62 & \textbf{72.87} & 71.78 & \textbf{71.74} & 70.82 & \textbf{79.68} & 78.13 & \textbf{89.28} & 87.86 \\
    5\%  & \textbf{59.43} & 58.29 & \textbf{72.82} & 72.48 & \textbf{70.51} & 70.19 & \textbf{79.40} & 78.61 & \textbf{88.48} & 87.87 \\
    10\% & \textbf{59.33} & 58.98 & \textbf{73.01} & 72.62 & \textbf{70.86} & 70.12 & \textbf{79.43} & 79.15 & \textbf{88.72} & 88.08 \\
    \bottomrule
    \end{tabular}
    }
\end{table*}

\textbf{Energy Efficiency and Compression Cost.}
A key motivation for prioritizing dataset selection over dataset synthesis is the prohibitive computational overhead of synthesis-based methods. As shown in Figure~\ref{fig_ablation_general_energy}~(b), we evaluate the end-to-end processing cost required to obtain a 3\% coreset on Flickr30K. Synthesis methods such as LoRS and RepBlend incur substantial GPU hours and energy consumption because they rely on continuous gradient matching, forward-backward passes, or complex synthetic features. In contrast, CAST adheres strictly to the dataset selection paradigm, directly extracting a representative coreset rather than synthesizing new samples. By bypassing generative hallucination and its associated computational burden, CAST provides an energy-efficient solution.




\subsection{Ablation Studies}

\begin{figure}
    \centering
    \includegraphics[width=1\linewidth]{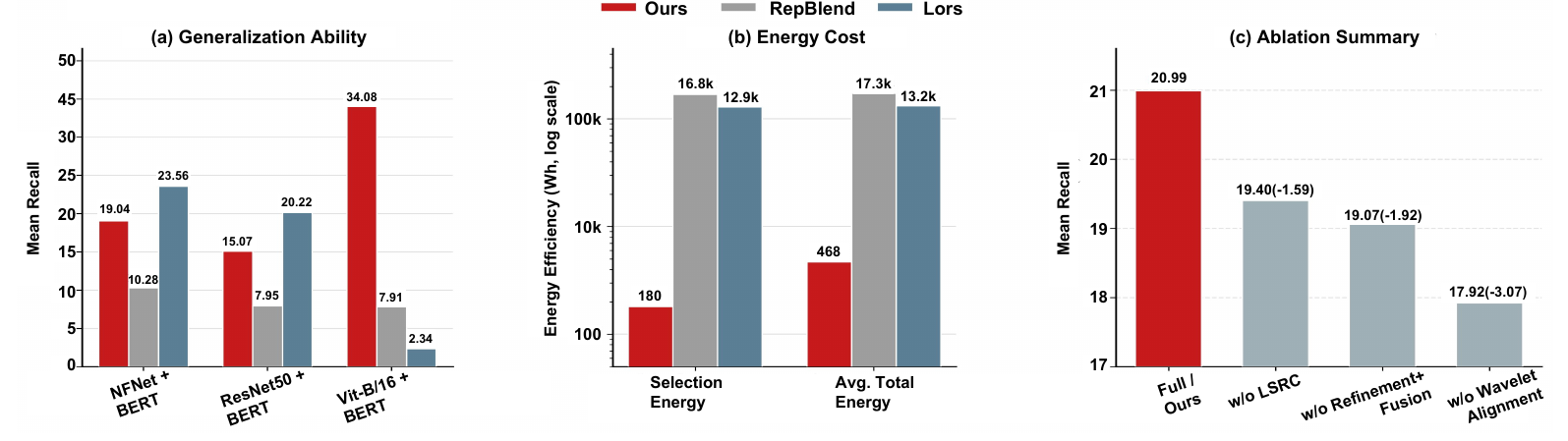}
    \caption{Summary of CAST (a) generalization, (b) efficiency, and (c) ablation results.}
    \label{fig_ablation_general_energy}
\end{figure}

\textbf{Component Ablation.}
We conduct component-wise ablations on Flickr30K under a constrained 3\% compression budget. The full CAST model is compared with three variants: \textbf{w/o LSRC}, which removes Local Soft Relational Coverage; \textbf{w/o Refinement \& Fusion}, which bypasses local-collapse-aware refinement and adaptive cross-modal fusion; and \textbf{w/o Wavelet Alignment}, which replaces multi-scale diffusion wavelet matching with standard frequency-domain alignment. As shown in Figure~\ref{fig_ablation_general_energy}~(c), all components are necessary. Removing Wavelet Alignment yields the largest Mean Recall drop ($-3.07$), highlighting the importance of multi-scale topology-aware distribution matching. Removing Refinement \& Fusion also degrades performance ($-1.92$), confirming the need to handle cross-modal structural imbalance. Without LSRC, performance drops by $1.59$ due to increased proxy redundancy. The full CAST achieves the best Mean Recall of $20.99$, demonstrating the complementarity of these modules. Additional ablations are provided in the Appendix~\ref{appendix_ablation_supplement}.

\textbf{Hyperparameter Sensitivity.}
We further analyze the sensitivity of CAST to the diffusion wavelet scale $s$. On Flickr30K under a 3\% budget, we compare single-scale settings ($s \in \{1,2,4\}$) with multi-scale spans $\{1,2,4\}$, $\{1,4,8\}$, and $\{1,8,16\}$. As shown in Table~\ref{tab_scale_ablation}, the results show that $\{1,2,4\}$ achieves the best average retrieval performance, confirming the benefit of jointly capturing local high-frequency boundaries and global low-frequency structures. In contrast, overly large spans cause severe degradation due to topological over-smoothing, which washes out discriminative semantic boundaries. Therefore, we adopt $\{1,2,4\}$ as the default configuration.


\begin{table*}[t]
    \centering
    \caption{Sensitivity analysis of different diffusion wavelet scales on bidirectional retrieval performance. The best results for each task are highlighted in \textbf{bold}.}
    \label{tab_scale_ablation}
    \resizebox{\textwidth}{!}{
    \begin{tabular}{l | ccc ccc | ccc ccc}
    \toprule
    \multirow{2}{*}{\textbf{Metric}} & \multicolumn{6}{c|}{\textbf{Image-to-Text}} & \multicolumn{6}{c}{\textbf{Text-to-Image}} \\
    \cmidrule{2-13}
    & \{1\} & \{2\} & \{4\} & \{1,2,4\} & \{2,4,8\} & \{4,8,16\} & \{1\} & \{2\} & \{4\} & \{1,2,4\} & \{2,4,8\} & \{4,8,16\} \\
    \midrule
    R@1  & \textbf{5.00} & 4.00 & 3.70 & 4.94 & 3.60 & 3.40 & 1.78 & 2.38 & 1.68 & \textbf{2.44} & 2.32 & 2.20 \\
    R@5  & 12.20 & 12.60 & 11.10 & \textbf{13.00} & 12.10 & 11.00 & 7.94 & 8.38 & 7.48 & \textbf{8.48} & 8.38 & 8.80 \\
    R@10 & 19.40 & 19.30 & 18.70 & \textbf{19.80} & 19.78 & 18.10 & 13.38 & 14.18 & 13.00 & \textbf{14.28} & 13.56 & 14.26 \\
    \bottomrule
    \end{tabular}
    }
\end{table*}



\section{Conclusion}

In this paper, we propose CAST, a novel \textbf{C}ollapse-\textbf{A}ware multi-\textbf{S}cale \textbf{T}opology fusion framework to address the prohibitive computational overhead of training large multimodal models. To resolve fine-grained cross-modal information imbalance, we integrate modality-specific graphs via local-collapse-aware refinement and adaptive cross-modal fusion. Furthermore, to capture location-dependent frequencies, we formulate a multi-scale distribution matching criterion in the wavelet domain, preserving both global structural skeletons and local high-frequency semantics. Additionally, our local soft relational coverage mechanism effectively mitigates proxy redundancy in dense regions, maximizing coreset diversity. Extensive evaluations demonstrate that CAST outperforms state-of-the-art selection baselines, achieving cross-architecture generalization and energy efficiency comparable to computationally expensive synthesis methods. Ultimately, CAST provides a robust and sustainable paradigm for advancing multimodal models under strict computational constraints.

\bibliographystyle{unsrtnat}
\bibliography{ref}


\newpage

\appendix

\section{Technical appendices and supplementary material}


\subsection{Details of Unified Graph Reconstruction}
\label{appendix_fusion}

As described in the main text, after deriving the response entropy $H_s^{(m)}$, a lower entropy implies that the structural representation at scale $s$ is heavily concentrated on localized regions, indicating a potential risk of structural collapse. To explicitly quantify this, we define the scale collapse score as:
$$C_s^{(m)}=1-H_s^{(m)}.$$
Based on this metric, we map the bimodal collapse scores into adaptive fusion weights using a softmax function parameterized by a temperature $T$:$$\gamma_s^{(m)}=\frac{\exp(-C_s^{(m)}/T)}{\exp(-C_s^{(I)}/T)+\exp(-C_s^{(T)}/T)}.$$
This formulation ensures that the modality with a lower degree of collapse at a specific scale is assigned a higher weight. Consequently, the target unified response at scale $s$ is constructed by linearly combining the bimodal responses:
$$Y_s^{\text{target}}=\gamma_s^{(I)}W_s^{(I)}+\gamma_s^{(T)}W_s^{(T)}.$$

Subsequently, rather than naively averaging in the original edge-weight space, we inversely reconstruct the unified graph over the joint candidate edge support set of the refined bimodal graphs. For each candidate edge $(i,j)$, the unified edge weight is estimated based on the cosine similarity of the node pairs within the multi-scale target consensus response: $$\tilde{B}_{ij}^{*}=\sum_{s\in\mathcal{S}}\omega_s\max\left(0,\cos\bigl(Y_{s,i}^{\text{target}},Y_{s,j}^{\text{target}}\bigr)\right),$$
where $\omega_s$ represents the predefined scale coefficient.

Finally, to guarantee that the unified graph imposes valid and reasonable structural constraints, we apply non-negative, symmetric, and sparsity-inducing operations:$$B_{ij}^{*}=\operatorname{Sym}\left(\max(\tilde{B}_{ij}^{*}-\lambda_{\text{sp}},0)\right),$$where $\lambda_{\text{sp}}$ is the threshold controlling the graph sparsity. Through this wavelet-domain multi-scale fusion and inverse reconstruction, the resulting unified latent topology $B^*$ effectively absorbs both the local texture boundaries from the image modality and the global semantic skeletons from the text modality, laying a robust geometric foundation for the subsequent coreset sampling.

\subsection{Details of Soft Relation Graph Construction and Discrete Realization}
\label{appendix_LSRC_and_matching}

\textbf{Construction of the Soft Relation Graph $R$.}
As introduced in the main text, the Local Soft Relational Coverage (LSRC) module mitigates redundant proxy assignments by propagating the direct spatial coverage $h_i$ into a relation-aware indirect coverage $\bar{h}_i$. This propagation relies on the construction of the local soft relation graph $R$.
Specifically, for any two authentic nodes $z_i$ and $z_j$ in the spatial domain, we first define a base geometric decay term:
$$g_{ij} = \exp\left(-\frac{\|z_i-z_j\|_2^2}{\sigma^2}\right)$$
where $\sigma$ controls the spatial bandwidth. To incorporate the underlying topological priors rather than relying solely on Euclidean distances, we fuse this geometric decay with the local cross-modal topological support $r_{ij}^{\text{cross}}$ derived from the refined bimodal graphs. The soft relation graph $R$ is thus formulated as:
$$R_{ij} = g_{ij}\cdot\left[\eta\cdot r_{ij}^{\text{cross}}+(1-\eta)\right]$$
where $\eta \in [0,1]$ is a balancing hyperparameter.

Subsequently, the direct spatial coverage $h_i$ of the proxy points $Y=\{y_1, \dots, y_K\}$ on node $z_i$ is explicitly calculated as:
$$h_i = \frac{1}{K}\sum_{k=1}^{K} \exp\left(-\frac{\|z_i-y_k\|_2^2}{\tau_c}\right)$$
where $\tau_c$ is the coverage temperature. Operating like a diffusion process on the relation network, this direct coverage is propagated via $R$ to form the relation-aware indirect coverage $\bar{h}_i$:
$$\bar{h}_i = (1-\beta)h_i + \beta\sum_j R_{ij}h_j$$
where $\beta$ denotes the propagation rate. This mechanism ensures that a proxy point structurally represents not only its immediate geometric neighbors but also the broader local semantic cluster, laying the foundation for the $L_{\text{LSRC}}$ formulation in the main text.

\textbf{Cost Formulation for Discrete Realization.}
Upon convergence of the continuous proxies, we map them back to the authentic dataset to form the discrete coreset $\mathcal{C}$. Following the protocol established in FAST, for each proxy point $y_k$ and authentic node $z_i$, we construct a comprehensive matching cost matrix $C_{k,i}$:
$$C_{k,i} = \alpha_d d_{\text{diff}}(y_k,z_i) + \alpha_w d_{\text{wavelet}}(y_k,z_i) + \alpha_t c_{\text{topo}}(i) - \alpha_q q_i$$
Here, $d_{\text{diff}}$ evaluates the geometric distance in the unified spatial domain, $d_{\text{wavelet}}$ measures the discrepancy in the multi-scale wavelet signature space, and $c_{\text{topo}}(i)$ represents the intrinsic topological cost of the authentic node. Notably, $q_i$ denotes the relation-aware coverage confidence derived from LSRC. The negative sign preceding $\alpha_q$ intrinsically rewards nodes that possess higher representativeness within their local relational coverage.

Finally, we apply the Hungarian matching algorithm to solve for the optimal one-to-one discrete assignment:
$$\pi^* = \arg\min_{\pi} \sum_{k=1}^{K} C_{k,\pi(k)}$$
The authentic image-text pairs corresponding to the optimized indices $\{\pi^*(1), \dots, \pi^*(K)\}$ constitute the final discrete coreset $\mathcal{C}$.

\begin{table}[h]
    \centering
    \caption{Detailed ablation study on Flickr across different budget values.}
    \label{tab:ablation_flickr_appendix}
    \vspace{-8pt}
    \resizebox{\textwidth}{!}{
    \begin{tabular}{lrrrrrrr}
        \toprule
        \textbf{Method} & \textbf{Budget} & \textbf{TR@1} & \textbf{TR@5} & \textbf{TR@10} & \textbf{IR@1} & \textbf{IR@5} & \textbf{IR@10} \\
        \midrule
        Full / Ours & 100 & 3.9 & 10.8 & 16.7 & 1.50 & 6.24 & 10.86 \\
        Full / Ours & 200 & 4.4 & 13.0 & 19.8 & 2.24 & 8.38 & 14.18 \\
        Full / Ours & 500 & 7.5 & 19.3 & 28.0 & 3.56 & 12.08 & 20.16 \\
        Full / Ours & 0.01 & 7.5 & 21.7 & 31.1 & 4.54 & 16.66 & 25.28 \\
        Full / Ours & 0.02 & 7.6 & 23.5 & 33.7 & 5.12 & 18.32 & 27.36 \\
        Full / Ours & 0.03 & 7.9 & 23.9 & 34.6 & 6.54 & 21.62 & 31.38 \\
        \midrule
        w/o LSRC & 100 & 2.5 & 9.3 & 16.6 & 1.42 & 6.02 & 9.98 \\
        w/o LSRC & 200 & 3.3 & 11.7 & 18.8 & 1.84 & 8.04 & 13.68 \\
        w/o LSRC & 500 & 6.9 & 18.9 & 28.7 & 3.94 & 12.34 & 20.06 \\
        w/o LSRC & 0.01 & 5.1 & 19.7 & 30.4 & 4.06 & 15.34 & 24.20 \\
        w/o LSRC & 0.02 & 7.1 & 21.9 & 32.8 & 5.32 & 17.78 & 27.48 \\
        w/o LSRC & 0.03 & 6.4 & 22.5 & 32.9 & 6.24 & 19.26 & 29.08 \\
        \midrule
        w/o Correction + Adaptive Fusion & 100 & 3.1 & 8.7 & 14.7 & 1.26 & 4.96 & 9.26 \\
        w/o Correction + Adaptive Fusion & 200 & 4.8 & 12.8 & 18.6 & 1.82 & 7.34 & 13.38 \\
        w/o Correction + Adaptive Fusion & 500 & 5.0 & 17.5 & 26.4 & 3.02 & 11.44 & 18.76 \\
        w/o Correction + Adaptive Fusion & 0.01 & 5.3 & 19.3 & 29.6 & 4.32 & 14.70 & 22.94 \\
        w/o Correction + Adaptive Fusion & 0.02 & 6.4 & 20.4 & 29.9 & 4.36 & 16.78 & 25.94 \\
        w/o Correction + Adaptive Fusion & 0.03 & 6.4 & 20.9 & 32.8 & 5.70 & 19.34 & 29.26 \\
        \midrule
        w/o Wavelet Alignment & 100 & 2.1 & 9.5 & 15.6 & 1.48 & 6.02 & 9.98 \\
        w/o Wavelet Alignment & 200 & 3.0 & 10.4 & 17.7 & 1.76 & 6.98 & 11.74 \\
        w/o Wavelet Alignment & 500 & 4.9 & 17.3 & 25.9 & 3.06 & 11.26 & 18.28 \\
        w/o Wavelet Alignment & 0.01 & 5.9 & 18.1 & 29.1 & 4.64 & 14.78 & 23.34 \\
        w/o Wavelet Alignment & 0.02 & 8.0 & 22.3 & 31.7 & 5.16 & 16.38 & 25.50 \\
        w/o Wavelet Alignment & 0.03 & 6.4 & 19.6 & 30.8 & 5.48 & 18.00 & 27.26 \\
        \bottomrule
    \end{tabular}
    }
\end{table}

\subsection{Ablation Supplement}
\label{appendix_ablation_supplement}
Table~\ref{tab:ablation_flickr_appendix} reports the detailed ablation results on Flickr under different budget values. The results further validate the contribution of each component in CAST, including local-collapse-aware refinement with adaptive fusion, LSRC, and wavelet alignment.

\subsection{Results on the MS-COCO Dataset}
\label{appendix_COCO_result}

Table~\ref{tab_coco} reports the bidirectional image-text retrieval results on the COCO dataset under extremely low compression budgets. Overall, CAST achieves the best performance on most retrieval metrics, obtaining the top result on 15 out of 18 metric-budget combinations. In particular, CAST consistently outperforms all unimodal and multimodal selection baselines on all text retrieval metrics, including TR@1, TR@5, and TR@10, across the three budgets. This demonstrates that the proposed collapse-aware topology refinement and multi-scale matching strategy effectively preserves cross-modal semantic information for text-to-image retrieval.

\begin{table*}[t]
    \centering
    \caption{Results of bidirectional image-text retrieval on the COCO dataset across varying compression budgets. The best performance is highlighted in \textbf{bold}.}
    \label{tab_coco}
    \resizebox{\textwidth}{!}{
    \setlength{\tabcolsep}{3pt}
    \begin{tabular}{cc | ccccccc | cccc | c}
    \toprule
    \multirow{2}{*}{\textbf{Budget}} & \multirow{2}{*}{\textbf{Metric}} 
    & \multicolumn{7}{c|}{\textbf{Unimodal Selection Baselines}} 
    & \multicolumn{4}{c|}{\textbf{Multimodal Selection Baselines}} 
    & \multirow{2}{*}{\textbf{Ours}} \\
    \cmidrule{3-13}
    & & Random & Herding & Glister & EL2N & Entropy & GradMatch & FAST 
      & ViSA & PreSel & Dyn-Prune & Prophet & \\
    \midrule

    \multirow{6}{*}{100} 
    & IR@1  & 0.43 & 0.88 & 0.84 & 0.30 & 0.44 & 0.56 & \textbf{0.91} & 0.35 & 0.42 & 0.62 & 0.78 & 0.84 \\
    & IR@5  & 1.61 & 3.38 & 3.52 & 1.44 & 1.78 & 2.90 & 3.61 & 1.36 & 1.58 & 2.36 & 3.04 & \textbf{3.74} \\
    & IR@10 & 3.05 & 6.04 & 6.48 & 2.52 & 3.12 & 5.58 & 6.42 & 2.72 & 3.08 & 4.92 & 5.84 & \textbf{6.51} \\
    & TR@1  & 0.90 & 0.37 & 0.45 & 0.17 & 0.26 & 0.41 & 0.72 & 0.31 & 0.34 & 0.53 & 0.68 & \textbf{1.00} \\
    & TR@5  & 2.94 & 1.66 & 1.82 & 0.80 & 1.28 & 1.82 & 3.48 & 1.54 & 1.76 & 2.38 & 2.71 & \textbf{4.26} \\
    & TR@10 & 5.40 & 3.14 & 3.57 & 1.64 & 2.38 & 3.32 & 6.12 & 2.93 & 3.16 & 4.38 & 4.92 & \textbf{7.70} \\

    \midrule

    \multirow{6}{*}{200} 
    & IR@1  & 0.64 & 1.32 & 0.88 & 0.40 & 0.58 & 1.30 & \textbf{1.42} & 0.58 & 0.67 & 0.92 & 1.18 & 1.36 \\
    & IR@5  & 3.06 & 4.60 & 3.82 & 1.96 & 2.70 & 4.44 & 4.54 & 2.48 & 2.76 & 3.82 & 4.61 & \textbf{4.78} \\
    & IR@10 & 5.65 & 8.12 & 6.76 & 3.40 & 4.44 & 7.80 & 8.72 & 4.84 & 5.12 & 6.92 & 7.86 & \textbf{8.79} \\
    & TR@1  & 1.18 & 0.56 & 0.56 & 0.26 & 0.38 & 0.64 & 1.21 & 0.48 & 0.53 & 0.72 & 0.96 & \textbf{1.94} \\
    & TR@5  & 5.24 & 2.59 & 2.45 & 1.16 & 1.70 & 2.73 & 5.68 & 2.26 & 2.57 & 3.62 & 4.74 & \textbf{6.36} \\
    & TR@10 & 9.24 & 4.79 & 4.55 & 2.26 & 3.21 & 4.88 & 9.86 & 4.28 & 4.75 & 6.42 & 7.82 & \textbf{11.26} \\

    \midrule

    \multirow{6}{*}{500} 
    & IR@1  & 1.29 & 2.44 & 1.70 & 0.70 & 1.16 & 2.46 & 2.58 & 1.18 & 1.32 & 1.74 & 2.15 & \textbf{2.72} \\
    & IR@5  & 5.30 & 7.58 & 6.84 & 3.36 & 4.88 & 7.30 & \textbf{8.34} & 4.26 & 4.72 & 6.28 & 7.46 & 8.24 \\
    & IR@10 & 9.29 & 14.58 & 11.28 & 5.78 & 8.20 & 11.96 & 14.32 & 7.66 & 8.35 & 10.84 & 12.73 & \textbf{14.66} \\
    & TR@1  & 2.28 & 1.26 & 1.16 & 0.46 & 0.82 & 1.11 & 2.36 & 0.92 & 1.06 & 1.42 & 1.78 & \textbf{2.58} \\
    & TR@5  & 8.44 & 4.98 & 4.39 & 2.02 & 3.24 & 4.58 & 8.76 & 3.86 & 4.24 & 5.82 & 7.16 & \textbf{9.10} \\
    & TR@10 & 14.22 & 8.86 & 7.73 & 3.85 & 5.85 & 7.99 & 14.68 & 7.18 & 7.86 & 10.24 & 12.58 & \textbf{14.96} \\

    \bottomrule
    \end{tabular}
    }
\end{table*}

\subsection{Cross-architecture Results of Synthesis-based Methods}
\label{appendix_synthesis_cross_architecture}

Table~\ref{tab:synthesis_cross_architecture} reports the cross-architecture retrieval results of two synthesis-based methods, RepBlend and LORS, when ResNet-10 is used as the upstream model. We evaluate the synthesized data on multiple downstream backbones, including NFNet, ResNet-50, ViT-B/16, and ResNet-10. The results show that the performance of synthesis-based methods varies significantly across evaluation backbones, indicating that synthetic data generated from a fixed upstream architecture may not consistently transfer to different downstream architectures.

\begin{table}[t]
\centering
\caption{Cross-architecture results of synthesis-based methods using ResNet-10 as the upstream model.}
\label{tab:synthesis_cross_architecture}
\vspace{-6pt}
\resizebox{0.8\linewidth}{!}{
\begin{tabular}{llrrrrrr}
\toprule
\textbf{Method} & \textbf{Eval. Backbone} & \textbf{TR@1} & \textbf{TR@5} & \textbf{TR@10} & \textbf{IR@1} & \textbf{IR@5} & \textbf{IR@10} \\
\midrule
RepBlend & NFNet     & 2.40 & 6.40 & 11.90 & 0.86 & 3.54 & 6.34 \\
RepBlend & ResNet-50 & 0.20 & 1.00 & 2.00 & 0.22 & 1.00 & 1.92 \\
RepBlend & ViT-B/16  & 1.66 & 6.84 & 11.28 & 2.30 & 9.10 & 16.30 \\
RepBlend & ResNet-10 & 0.30 & 2.30 & 4.10 & 0.40 & 2.16 & 3.76 \\
\midrule
LORS & NFNet     & 0.10 & 0.50 & 1.00 & 0.16 & 0.62 & 1.14 \\
LORS & ResNet-50 & 0.70 & 3.50 & 5.30 & 0.72 & 3.02 & 5.10 \\
LORS & ViT-B/16  & 0.10 & 0.40 & 1.10 & 0.10 & 0.70 & 1.32 \\
LORS & ResNet-10 & 0.20 & 0.70 & 1.70 & 0.24 & 0.98 & 1.68 \\
\midrule
Ours & NFNet     & 7.70 & 22.30 & 32.00 & 5.48 & 18.74 & 28.00 \\
Ours & ResNet-50 & 5.00 & 16.00 & 24.40 & 4.90 & 15.68 & 24.44 \\
Ours & ViT-B/16  & 16.80 & 43.30 & 55.00 & 11.74 & 32.50 & 45.12 \\
Ours & ResNet-10 & 3.80 & 12.60 & 19.40 & 3.24 & 10.82 & 17.36 \\
\bottomrule
\end{tabular}
}
\end{table}

\subsection{Flickr30K Results under Ratio-based Budgets}
\label{appendix_flickr_ratio_results}

Table~\ref{tab_flickr_ratio} reports the bidirectional image-text retrieval results on the Flickr30K dataset under ratio-based compression budgets of 1\%, 2\%, and 3\%. We compare CAST against representative unimodal selection baselines (e.g., Random, Herding, Glister, EL2N, Entropy, GradMatch, and FAST), as well as multimodal selection baselines (e.g., ViSA, PreSel, Dynamic Pruning, and Data Prophet). The results show that CAST consistently achieves strong overall performance across different retrieval directions and budget ratios.

\begin{table*}[t]
    \centering
    \caption{Results of bidirectional image-text retrieval on the Flickr30K dataset across ratio-based compression budgets. The best performance is highlighted in \textbf{bold}.}
    \label{tab_flickr_ratio}
    \resizebox{\textwidth}{!}{
    \setlength{\tabcolsep}{3pt}
    \begin{tabular}{cc | ccccccc | cccc | c}
    \toprule
    \multirow{2}{*}{\textbf{Budget}} & \multirow{2}{*}{\textbf{Metric}} 
    & \multicolumn{7}{c|}{\textbf{Unimodal Selection Baselines}} 
    & \multicolumn{4}{c|}{\textbf{Multimodal Selection Baselines}} 
    & \multirow{2}{*}{\textbf{Ours}} \\
    \cmidrule{3-13}
    & & Random & Herding & Glister & EL2N & Entropy & GradMatch & FAST 
      & ViSA & PreSel & Dyn-Prune & Prophet & \\
    \midrule

    \multirow{6}{*}{0.01} 
    & IR@1  & 3.64 & 1.76 & 1.76 & 1.46 & 1.18 & 1.64 & 4.18 & 3.14 & 2.48 & 2.24 & 2.98 & \textbf{4.54} \\
    & IR@5  & 13.30 & 7.04 & 6.38 & 6.56 & 4.24 & 6.50 & 15.14 & 9.74 & 9.58 & 8.74 & 10.22 & \textbf{16.66} \\
    & IR@10 & 21.62 & 12.62 & 11.96 & 10.48 & 7.22 & 11.20 & 23.78 & 15.12 & 14.24 & 14.24 & 15.82 & \textbf{25.28} \\
    & TR@1  & 6.10 & 2.20 & 2.10 & 1.80 & 0.50 & 2.20 & 6.60 & 3.70 & 4.40 & 3.40 & 4.00 & \textbf{7.50} \\
    & TR@5  & 20.30 & 8.30 & 8.40 & 6.80 & 3.00 & 7.90 & 20.20 & 11.70 & 11.40 & 10.30 & 11.50 & \textbf{21.70} \\
    & TR@10 & 30.00 & 14.60 & 14.10 & 11.80 & 5.00 & 12.70 & 30.50 & 18.20 & 17.70 & 16.50 & 17.40 & \textbf{31.10} \\

    \midrule

    \multirow{6}{*}{0.02} 
    & IR@1  & 5.24 & 5.34 & 5.24 & 4.80 & 3.92 & 4.92 & \textbf{5.50} & 3.04 & 3.58 & 2.36 & 3.48 & 5.42 \\
    & IR@5  & 15.12 & 17.50 & 17.04 & 14.74 & 12.34 & 15.92 & 17.46 & 11.64 & 11.60 & 9.14 & 12.60 & \textbf{18.32} \\
    & IR@10 & 17.93 & 26.28 & 25.84 & 22.10 & 18.86 & 24.64 & 26.88 & 17.60 & 18.00 & 15.42 & 18.56 & \textbf{27.36} \\
    & TR@1  & 6.00 & 6.90 & 6.80 & 5.80 & 3.20 & 7.50 & 6.70 & 4.40 & 3.70 & 3.00 & 5.30 & \textbf{7.60} \\
    & TR@5  & 22.60 & 20.80 & 20.90 & 18.00 & 12.40 & 19.60 & 22.10 & 11.50 & 11.90 & 11.50 & 13.90 & \textbf{23.50} \\
    & TR@10 & 32.00 & 29.80 & 32.10 & 26.40 & 18.20 & 30.30 & 33.20 & 17.10 & 17.40 & 18.60 & 19.20 & \textbf{33.70} \\

    \midrule

    \multirow{6}{*}{0.03} 
    & IR@1  & 5.86 & 6.06 & 5.86 & 4.74 & 4.82 & 5.54 & 6.50 & 4.26 & 4.32 & 3.36 & 5.04 & \textbf{6.54} \\
    & IR@5  & 19.16 & 18.86 & 17.76 & 17.24 & 14.98 & 17.18 & 20.58 & 13.44 & 12.52 & 11.24 & 13.60 & \textbf{21.62} \\
    & IR@10 & 28.64 & 28.58 & 27.52 & 25.54 & 22.14 & 27.24 & 29.88 & 19.64 & 18.88 & 17.94 & 20.68 & \textbf{31.38} \\
    & TR@1  & 7.70 & \textbf{8.42} & 5.90 & 7.80 & 6.40 & 7.30 & 8.40 & 4.20 & 3.80 & 3.80 & 5.70 & 8.27 \\
    & TR@5  & 21.00 & 20.60 & 19.50 & 20.20 & 15.20 & 21.60 & 22.60 & 14.70 & 10.90 & 11.60 & 15.00 & \textbf{23.90} \\
    & TR@10 & 32.00 & 33.00 & 30.00 & 29.70 & 22.70 & 33.10 & 33.70 & 22.00 & 16.30 & 18.50 & 21.40 & \textbf{34.60} \\

    \bottomrule
    \end{tabular}
    }
\end{table*}
\subsection{Limitations}
\label{appendix_limitations}

While CAST shows promising performance in multimodal coreset selection, there remain several directions for further exploration.
First, our current study mainly evaluates CAST on image-text retrieval benchmarks. 
Although these benchmarks provide a standard and representative setting for multimodal selection, extending the evaluation to broader multimodal scenarios, such as instruction tuning or video-language tasks, would further demonstrate its generality.
Second, CAST currently uses fixed hyperparameter settings across experiments, including neighborhood size, diffusion scales, and loss weights. 
Exploring adaptive parameter selection may further improve convenience and robustness when applying CAST to new datasets.
Third, our implementation focuses on validating the effectiveness of topology-aware selection, rather than optimizing system-level efficiency. 
Future work could further accelerate graph construction and multi-scale matching for larger-scale applications.

\subsection{Broader Impacts}
\label{appendix_broader_impacts}

CAST aims to reduce the computational and energy costs of training multimodal models by selecting compact and informative coresets, which may improve the accessibility and sustainability of multimodal learning.
Since the selected coresets are derived from existing multimodal datasets, their quality and safety are still influenced by the underlying data sources.
Therefore, selected coresets should be used together with standard data governance practices, including dataset license checks, privacy filtering, bias evaluation, and safety screening, especially before real-world deployment.
Overall, CAST is intended to support more efficient and responsible multimodal learning.
\clearpage

\end{document}